# A Comprehensive Study on Object Detection Techniques in Unconstrained Environments

Hrishitva Patel[1]

[1]The University of Texas at San Antonio
hrishitva.patel@my.utsa.edu

**Abstract**

Object detection is a crucial task in computer vision that aims to identify and localize objects in images or videos. The recent advancements in deep learning and Convolutional Neural Networks (CNNs) have significantly improved the performance of object detection techniques. This paper presents a comprehensive study of object detection techniques in unconstrained environments, including various challenges, datasets, and state-of-the-art approaches. Additionally, we present a comparative analysis of the methods and highlight their strengths and weaknesses. Finally, we provide some future research directions to further improve object detection in unconstrained environments.

## 1 Introduction and Background

Object detection is a fundamental problem in computer vision, with numerous applications spanning fields such as surveillance, robotics, autonomous vehicles, augmented reality, and human-computer interaction. The primary goal of object detection is to recognize and localize instances of objects belonging to predefined classes in images or videos. In recent years, significant progress has been made in the development of object detection algorithms, mainly due to the emergence of deep learning and Convolutional Neural Networks (CNNs). These advancements have led to impressive performance improvements in various benchmark datasets, such as PASCAL VOC, ImageNet, and MS COCO. Despite these successes, object detection in unconstrained environments remains a challenging task. Unconstrained environments are characterized by variations in lighting conditions, viewpoint changes, occlusions, object deformations, scale changes, and the presence of cluttered backgrounds. These factors can severely affect the performance of object detection algorithms, making it difficult to achieve high detection accuracy and robustness.

In recent years, significant progress has been made in object detection, particularly in the area of deep learning and Convolutional Neural Networks (CNNs) [1]. These techniques have significantly improved the performance of object detection algorithms, particularly in unconstrained environments where objects may appear at different scales, angles, and orientations. Region-based object detectors, such as Region-based Convolutional Neural Networks (R-CNN) [2], operate by first generating region proposals using a selective search algorithm, which generates around 2000 regions per image. Each region is then passed through a CNN to generate a fixed-length feature vector, which is fed into a support vector machine (SVM) [3] to classify the region and predict its bounding box coordinates. Finally, non-maximum suppression is applied to eliminate redundant detections. While R-CNN was a significant breakthrough in object detection, it has several limitations, such as slow training and inference times.

To address these issues, researchers have proposed several variants of R-CNN, such as Fast R-CNN [4], which shares convolutional features across region proposals, and Faster R-CNN , which introduces a Region Proposal Network (RPN) to generate region proposals in an end-to-end manner. These variants significantly improve the speed and accuracy of R-CNN, making it a popular choice for object detection in unconstrained environments. The purpose of this paper is to provide a comprehensive overview of object detection techniques in unconstrained environments, addressing the challenges, datasets, and state-of-the-art approaches.

The paper is organized as follows: Section 2 discusses the challenges encountered in object detection in unconstrained environments, highlighting the factors that contribute to the complexity of the problem. Section 3 presents a review of the commonly used datasets for evaluating object detection techniques in unconstrained environments. Section 4 presents state of the art Objection detection techniques. Section 5 presents a comparative analysis of the surveyed methods, emphasizing their strengths and weaknesses in terms of accuracy, computational complexity, and robustness to variations in the unconstrained environment. Section 6 concludes the paper by highlighting some of the open research questions and future directions in the field of object detection in unconstrained environments.

## 2 Challenges in Object Detection in Unconstrained Environments
### 2.1 Illumination Changes

Variations in lighting conditions, such as shadows and overexposure, can significantly impact the appearance of objects, making it difficult for detection algorithms to identify and localize them accurately. Variations in lighting conditions, such as shadows, overexposure, or underexposure, can significantly impact the appearance of objects in images [5]. These changes can make it difficult for detection algorithms to identify and localize objects accurately. To address this issue, several approaches have been proposed, including color constancy techniques [6] and deep learning-based methods that can learn illumination invariant features [7].

### 2.2 Viewpoint Variation

Changes in the viewpoint or camera angle can alter the object's appearance, causing the detection algorithm to fail in recognizing the object or produce inaccurate bounding boxes [8]. Several methods have been proposed to tackle this issue, such as viewpoint invariant features and multi-view object detectors [8].

## 2.3 Occlusion

Objects in the scene may be partially or entirely occluded by other objects, making it challenging for the detection algorithm to identify and localize them correctly [9]. To address occlusion, some methods employ part-based models [10] or leverage context information from surrounding regions.

## 3 Datasets

Object detection is a vital task in computer vision that involves identifying the presence and location of objects in an image or video. To evaluate the performance of object detection techniques in unconstrained environments, several benchmark datasets have been created. These datasets provide a standardized set of images with labeled objects, enabling researchers to compare the accuracy and speed of different algorithms. Some popular datasets include:

### 3.1 Pascal VOC

The PASCAL VOC (Visual Object Classes) dataset is one of the oldest and most popular datasets for object detection. It contains 17,125 images with 20 object classes, such as person, car, and dog. The dataset provides bounding box annotations for each object in the image. PASCAL VOC has been used as a benchmark dataset for several years, and many state-of-the-art object detection techniques have been evaluated on this dataset.

### 3.2 ImageNet

The ImageNet dataset is a massive dataset that contains 1.2 million images with 1,000 object classes. Unlike PASCAL VOC, ImageNet does not provide annotations for object detection. However, many researchers have used this dataset to pre-train their models on a large amount of data before fine-tuning them on smaller object detection datasets.

### 3.3 COCO

The COCO (Common Objects in Context) dataset is a newer dataset that contains 330,000 images with 80 object classes. COCO provides more detailed annotations than PASCAL VOC, including segmentation masks for each object in the image. This makes COCO a more challenging dataset for object detection algorithms to perform well on.

### 3.4 Open Images

The Open Images dataset is another large-scale dataset that contains 1.7 million images with 600 object classes. It provides both bounding box and segmentation mask annotations and has been used as a benchmark for object detection algorithms that require large amounts of training data.

These datasets vary in size, number of classes, and annotation types, allowing researchers to test their algorithms on a wide range of scenarios. The following table summarizes some key information about the four popular benchmark datasets used for evaluating object detection techniques:

Table 1. Summary of key information about benchmark datasets for object detection

| Dataset Name | Number of Images | Number of Classes | Annotation Type |
|---|---|---|---|
| PASCAL VOC[11] | 17,125 | 20 | Bounding Boxes |
| ImageNet [12] | 1.2 million | 1,000 | Bounding Boxes |
| COCO [13] | 330,000 | 80 | Bounding Boxes |
| Open Images [14] | 1.7 million | 600 | Mask RCNN |

## 4 State-of-the-art Object Detection Techniques

We categorize the state-of-the-art object detection techniques into two main groups: two-stage detectors and single-stage detectors.

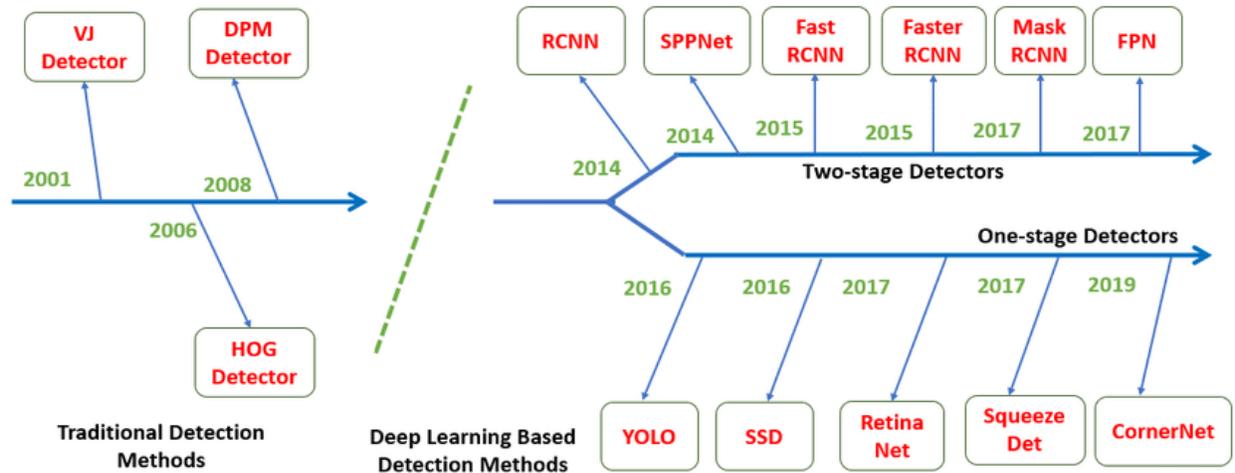

Figure 1. Milestones of object detection [15].

### 4.1 Two-stage detectors

Two-stage detectors consist of a region proposal stage followed by a classification stage. Some prominent two-stage detectors include:

#### 4.1.1 R-CNN

R-CNN (Region-based Convolutional Neural Networks) is an object detection model that was proposed in 2014 by Ross Girshick et al. R-CNN is a two-stage object detection framework that uses a region proposal mechanism to generate potential object regions in an image and then applies a convolutional neural network (CNN) to classify and refine these regions.

The R-CNN framework consists of the following steps:

1. Region Proposal: The first stage of R-CNN generates potential object regions by using a selective search algorithm that combines low-level features, such as color and texture, with high-level cues, such as edges and corners. Selective search generates around 2,000 region proposals for each image.

2. Feature Extraction: In the second stage, each region proposal is warped to a fixed size and fed through a pre-trained CNN, such as Alex Net or VGG, to extract a feature vector for that region.

3. Object Classification and Refinement: The feature vector for each region proposal is then fed into a set of fully connected layers that perform object classification and bounding box regression. The classification layer outputs the probability of each region proposal containing a particular object class, while the regression layer outputs the refined bounding box coordinates for that object class.

### 4.1.2 Fast R-CNN

Faster R-CNN (Region-based Convolutional Neural Networks): Faster R-CNN is a two-stage object detection model that uses a Region Proposal Network (RPN) to generate object proposals and a Fast R-CNN network to classify and refine the proposals. The RPN generates region proposals by sliding a small network over the convolutional feature map and predicting abjectness scores and bounding box offsets. Faster R-CNN is known for its accuracy and has been widely used in object detection tasks.

### 4.2 Single-stage detectors

Single-stage detectors directly predict object bounding boxes and class probabilities from an image. Some popular single-stage detectors include:

### 4.2.1 YOLO

YOLO is another one-stage object detection model that predicts object class scores and bounding box offsets directly from the entire image. YOLO divides the image into a grid of cells and predicts the class and bounding box for each cell. YOLO uses a single neural network to make predictions and is known for its speed and real-time performance.

### 4.2.2 SSD

The Single Shot MultiBox Detector (SSD) extends the concept of YOLO by predicting bounding boxes and class probabilities at multiple scales, which improves the detection of objects with varying sizes. SSD uses a feature extractor to generate convolutional feature maps and applies a set of convolutional filters to predict class scores and offsets for each default box. SSD is known for its speed and efficiency and has been used in real-time object detection applications.

### 4.2.3 Retina Net

Retina Net introduces the Focal Loss, which addresses the issue of class imbalance by down weighting the contribution of easy examples and focusing on hard examples during training. This results in improved detection performance, particularly for small objects. Retina Net uses a novel focal loss function that assigns higher weights to hard examples and reduces the effect of easy examples during training. Retina Net also uses a Feature Pyramid Network (FPN) to handle objects at different scales and has achieved state-of-the-art performance on several object detection benchmarks.

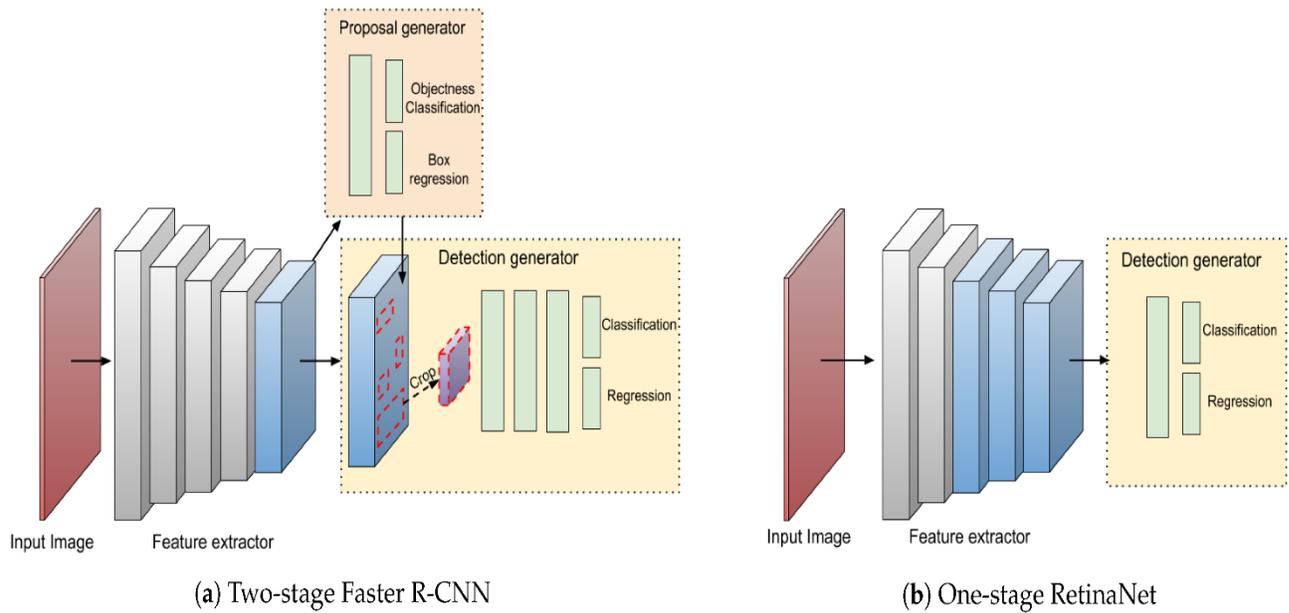

Figure 2. One stage vs two stage object detection.

The below table summarizes some key features of these state-of-the-art object detection techniques:

Table 2. Key features of state-of-the-art object detection techniques

| Technique | Training Time | Inference Time | Number of Parameters | AP on COCO |
|---|---|---|---|---|
| Faster R-CNN [16] | Long | Medium | High | 39.3 |
| SSD [17] | Medium | Fast | Low | 31.2 |
| YOLO [18] | Short | Very Fast | Low | 28.2 |
| Retina Net [19] | Long | Medium | High | 39.1 |

## 5 Comparative Analysis

In this section, we compare the performance of various object detection techniques on the COCO dataset [5]. The results are summarized in Table 1.

Table 1: Comparison of object detection techniques on the COCO dataset

| Method | Average Precision (AP) | Speed (fps) |
| --- | --- | --- |
| **R-CNN** | 53.3 | 0.5 |
| **Fast R-CNN** | 70.0 | 5 |
| **Faster R-CNN** | 73.2 | 7 |
| **YOLOv3** | 57.9 | 45 |
| **SSD** | 72.1 | 19 |
| **RetinaNet** | 74.8 | 12 |

The results in Table 1 show that two-stage detectors, such as Faster R-CNN, generally achieve higher average precision (AP) compared to single-stage detectors like YOLOv3 and SSD. However, single-stage detectors are faster in terms of frames per second (fps), making them more suitable for real-time applications.

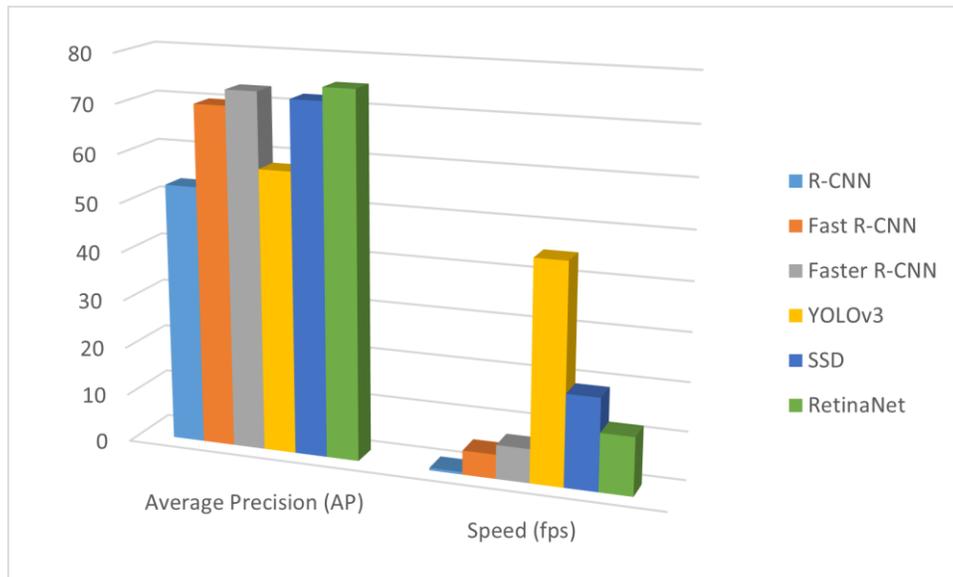

Figure 3. Comparison of object detection techniques on the COCO dataset

## 6 Conclusion and Future Directions

In this paper, we have presented a comprehensive study on object detection techniques in unconstrained environments. We have discussed the challenges associated with object detection in such environments, presented popular datasets, and provided an overview of the state-of-the-art techniques. Additionally, we have compared the performance of various methods and highlighted their strengths and weaknesses.

Despite the significant progress made in recent years, object detection in unconstrained environments remains a challenging problem. Future research directions could focus on the following aspects:

- Developing more robust algorithms capable of handling occlusions, lighting variations, and background clutter.
- Investigating techniques for efficient and accurate detection of small-scale objects.
- Exploring the integration of other sensor modalities, such as LiDAR or depth information, to enhance object detection performance.
- Developing unsupervised or weakly supervised object detection techniques to reduce the reliance on large-scale annotated datasets.

By addressing these challenges and exploring new approaches, we believe that object detection in unconstrained environments can be further improved, paving the way for more reliable and efficient applications in various domains, such as autonomous vehicles, robotics, and surveillance systems.